	\newif\ifboxes

	\documentclass[letterpaper, 10 pt, conference]{ieeeconf}  

	\IEEEoverridecommandlockouts                              
	
\newpage
\title{\LARGE \bf
	Non-Parametric Modeling of Spatio-Temporal Human Activity \mbox{Based on Mobile Robot Observations}
}

	\author{Marvin Stuede and Moritz Schappler$^{1}$
	\thanks{$^{1}$Authors are with the Leibniz University Hannover, Institute of Mechatronic Systems, D-30823 Garbsen, Germany,
	        {\tt\small marvin.stuede@imes.uni-hannover.de}}%
	}

	\newcommand{\capsize}{\fontsize{\small}\selectfont}
	\usepackage[compact]{titlesec}
	\usepackage[nolist]{acronym}
	\usepackage{cite}
	\usepackage{pdfpages}
	\usepackage{adjustbox}
	\usepackage{framed}
		\usepackage{array}
	\usepackage{multirow}
	\usepackage[ruled, linesnumbered]{algorithm2e}
	\usepackage{nicefrac}
	\usepackage{amsmath}
	
	\DeclareMathOperator*{\argmin}{arg\,min}
	\usepackage{amssymb}
	\usepackage{subcaption}
	\usepackage[font=small]{caption}
	\usepackage[T1]{fontenc}
	\usepackage[utf8]{inputenc}
	\usepackage{grffile}
	\usepackage{siunitx}
	\usepackage{booktabs}
	\usepackage{adjustbox}
	\usepackage{setspace}
	\usepackage{dblfloatfix} 
	\usepackage{etoolbox} 
	\usepackage[english]{babel}
	\usepackage{blindtext}
	\usepackage{pgfplots} 
		\usepackage{tikz}
		\usepgfplotslibrary[groupplots,dateplot]
	\usetikzlibrary{matrix}
	\usetikzlibrary{tikzmark}
	\usetikzlibrary{shadows,arrows.meta, shapes, fit, calc, positioning, backgrounds}
\pgfplotsset{compat=newest,
	every axis legend/.append style={at={(0.5,1.03)},anchor=south,nodes=right,font=\small}, 
	label style={font=\capsize} ,
	layers/my layer set/.define layer set={
		bg, 
		main,
		c1, 
		c2, 
		c3, 
	}{},
	set layers=my layer set,
}
	\pgfdeclarelayer{bg}    
\pgfdeclarelayer{c1}
\pgfdeclarelayer{c2}
\pgfdeclarelayer{c3}
\pgfsetlayers{bg, c1, c2, c3, main}

\usepackage{stmaryrd}
\usepackage{trimclip}

\makeatletter
\DeclareRobustCommand{\shortto}{%
	\mathrel{\mathpalette\short@to\relax}%
}

\newcommand{\short@to}[2]{%
	\mkern2mu
	\clipbox{{.5\width} 0 0 0}{$\m@th#1\vphantom{+}{\shortrightarrow}$}%
}
\makeatother

\definecolor{bt-blue}{RGB}{2,170,255}
	\newlength\picwidth
	\newlength\boxwidth
	\newlength\tikzwidth
	\newlength\figureheight
	\newlength\figurewidth
		\newlength\swidth
	\newlength\shorten
		\newif\ifcopyright
	\copyrighttrue

\makeatletter
\patchcmd{\@algocf@start}
{-1.5em}
{0pt}
{}{}
\makeatother
\makeatletter
\newcommand{\removelatexerror}{\let\@latex@error\@gobble}
\makeatother

\begin{document}

		\ifcopyright
					{\LARGE Notice}
		\newline
		\fboxrule=0.4pt \fboxsep=3pt
		
		\fbox{\begin{minipage}{1.1\linewidth}  
				Copyright (c) 2022 IEEE. Personal use of this material is permitted. For any other purposes, permission must be obtained from the IEEE by emailing pubs-permissions@ieee.org. \\
				
			Accepted to be published in: Proceedings of the 2022 IEEE/RSJ International Conference on Intelligent Robots and Systems (IROS), October 23 - 27, Kyoto, Japan 
				
		\end{minipage}}
		\else
		\fi

	\begin{acronym}
		\acro{bt}[BT]{Behavior Tree}
		\acro{psbt}[PSBT]{\textit{Person Search Behavior Tree}}
		\acro{nw}[NW]{\textit{never wait}}
		\acro{w}[W]{\textit{wait at the help location}}
		\acro{gc}[GC]{\textit{greedy planning to a close maximum}}
		\acro{gm}[GM]{\textit{greedy planning to the global maximum}}
		\acro{rnd}[RND]{\textit{Uniform random sampling of goals}}
		\acro{sbt}[SBT]{Stochastic Behavior Tree}
		\acro{gpr}[GPR]{Gaussian Process Regression}
		\acro{gp}[GP]{Gaussian Process}
		\acro{gmm}[GMM]{Gaussian Mixture Model}
		\acro{sgp}[SGP]{Sparse Gaussian Process}
		\acrodefplural{sgp}[SGPs]{Sparse Gaussian Processes}
		\acro{nudft}[NUDFT]{Non-uniform discrete Fourier transform}
		\acrodefplural{gp}[GPs]{Gaussian Processes}
		\acro{bts}[\textit{S}]{\textit{success}}
		\acro{btr}[\textit{R}]{\textit{running}}
		\acro{otsp}[OTSP]{open traveling salesman problem}
		\acro{btf}[\textit{F}]{\textit{failure}}
		\acro{sgd}[SGD]{Stochastic Gradient Descent}
		\acro{dtmc}[DTMC]{Discrete Time Markov Chain}
		\acro{mdp}[MDP]{Markov Decision Process}
		\acro{mcmc}[MCMC]{Markov Chain Monte Carlo}
		\acro{nrmse}[NRMSE]{normalized root mean square error}
		\acro{fov}[FOV]{field of view}
		\acro{rbf}[RBF]{radial basis function}
			\acro{nll}[NLL]{negative log-likelihood}
	\end{acronym}
	\newcommand*\prs[1][]{p_{\mathrm{s}#1}}
	\newcommand*\prf[1][]{p_{\mathrm{f}#1}}
	\newcommand{\wait}[1]{\mathcal{W}_{\mathrm{A},#1}}
	\newcommand{\place}[1]{\mathcal{P}_{#1}}
	\newcommand{\path}[2]{\mathcal{S}_{\mathrm{A},#1\to #2}}
	\newcommand{\waith}{\mathcal{W}_{\mathrm{A},\mathrm{h}}}
	\newcommand{\eal}{\textit{et al.}}
	\newcommand{\mvel}{\bar{v}}
	\newcommand{\lfail}{l_{\mathrm{fail}}}
	\newcommand{\ealcite}[1]{\eal~\cite{#1}}
	\newcommand{\bts}{\acs{bts}}
	\newcommand{\btr}{\acs{btr}}
	\newcommand{\btf}{\acs{btf}}
	\newcommand{\TODO}[1]{\textcolor{red}{\textbf{TODO: #1}}}
	\renewcommand{\vec}[1]{\mbox{\boldmath{$#1$}}}
	\newcommand{\dvec}[1]{\dot{\mbox{\boldmath{$#1$}}}}
	\newcommand{\ddvec}[1]{\ddot{\mbox{\boldmath{$#1$}}}}
	\newcommand{\ind}[1]{\mathrm{#1}}	
	\newcommand{\transpose}{\ind{T}}
	\newcommand{\ToDo}[1]{(\marginpar[\hfill!$\longrightarrow$]{\textcolor[rgb]{1,0,0}{\textbf{$\longleftarrow$!}}}\textcolor[rgb]{1,0,0}{\textbf{ToDo:}} \emph{#1})}
	\newcommand{\ToDoFloat}[1]{(\textcolor[rgb]{1,0,0}{\textbf{ToDo:}} \emph{#1})}
	\newcommand{\tmat}[2]{{^{\ind{#1}}\vec{T}_{\ind{#2}}}}
	\newcommand{\rmat}[2]{{^{\ind{#1}}\vec{R}_{\ind{#2}}}}
	\newcommand{\ks}[1]{{\ind{(KS)_{#1}}}}
	\maketitle
	\thispagestyle{empty}
	\pagestyle{empty}

	\newcommand{\Kxx}[2]{\rbf{K}_{#1#2}}
	\newcommand{\Kff}{\Kxx{\mathbf{f}}{\mkern 1mu\mathbf{f}}}
	\newcommand{\Kfuf}{\Kxx{\mathbf{f}}{\mathbf{u}_f}}
	\newcommand{\Kuff}{\Kxx{\mathbf{u}_f}{\mathbf{f}}}
	\newcommand{\Kufuf}{\Kxx{\mathbf{u}_f}{\mathbf{u}_f}}
	\newcommand{\rbf}[1]{\mathbf{#1}}
	\newcommand{\inv}{^{-1}}
	\newcommand{\xs}{\vec{x}_{\mathrm{s}}}
	\newcommand{\ls}{l_{\mathrm{s}}}
	\newcommand{\sigmas}{\sigma_{\mathrm{s}}}
	\newcommand{\xt}{x_{\mathrm{t}}}
	\newcommand{\domst}{\mathcal{S}_i}
	\newcommand{\doms}{\mathcal{T}}
	\newcommand{\nmax}{n_{\mathrm{max}}}
	\newcommand{\pmax}{\psi_{\mathrm{max}}}
	\newcommand*\diff{\mathop{}\!\mathrm{d}}

	\begin{abstract}
	This work presents a non-parametric spatio-temporal model for mapping human activity by mobile autonomous robots in a long-term context.
	Based on Variational Gaussian Process Regression, the model incorporates prior information of spatial and temporal-periodic dependencies to create a continuous representation of human occurrences.
	The inhomogeneous data distribution resulting from movements of the robot is included in the model via a heteroscedastic likelihood function and can be accounted for as predictive uncertainty.
	Using a sparse formulation, data sets over multiple weeks and several hundred square meters can be used for model creation.
	The experimental evaluation, based on multi-week data sets, demonstrates that the proposed approach outperforms the state of the art both in terms of predictive quality and subsequent path planning.
	
	\end{abstract}

	\section{Introduction}
\begin{figure}[t]
		
		\noindent
		\newcommand{\colarr}[1]{
			\node[inner sep=0pt,#1, anchor=south] (lower) {};
			\node[scale=1, above = 11mm of lower] (upper) {};
			\draw[line width=1pt, -{Latex[length=2mm,width=1.4mm]}] (lower.south) -- (upper.south);
		}
		\newcommand{\colbars}[1]{
			\colarr{above right= 0.75mm and #1mm of P.south west}
			\node[inner sep=0pt, align=center, right= \dbarlab of lower.south, anchor=north west, rotate=90] (c1)  {Observation dur.}; 
			\colarr{right= \cbarw of lower.south}
			\node[inner sep=0pt, align=center, right= \dbarlab of lower.south, anchor=north west, rotate=90] (c2) {Pred. mean}; 
			\colarr{right= \cbarw of lower.south}
			\node[inner sep=0pt, align=center, right= \dbarlab of lower.south, anchor=north west, rotate=90] (c3)  {Pred. variance}; 
		}
		\begin{center}
			\setlength{\figurewidth}{0.63\linewidth}
			\setlength{\figureheight}{0.7\figurewidth}

			\flushleft
							\begin{tikzpicture}
				\tikzset{
					header/.style={
						text width=.49\linewidth,
						align=center,
						inner sep=0pt,
						anchor=west
					}
				}
				\node[inner sep=1pt, anchor=south west, align=left] (P) at (0,0){\input{fig/hom_het.tex}};
			{\small 
									\newlength{\hgap}
					\setlength{\hgap}{0.1mm}
					\node[header, below right=1.0 mm and 0mm of P.south west, anchor=west] (b1) {(a)}; 
					\node[header, right=\hgap of b1.east, anchor=west] (b2) {(b)}; 
					\node[header, below=0.5mm of b1.south, anchor=north] (b3) {}; 
				}
				\end{tikzpicture}

%
				\begin{tikzpicture}
					\tikzset{
						header/.style={
							text width=.142\textwidth,
							align=center,
							inner sep=0pt,
							anchor=west
						}
					}
					\node[inner sep=0pt, anchor=south west, align=left] (P) at (0,0)
					{
					\includegraphics[trim=0 0 0 0.4, clip, width=\linewidth]{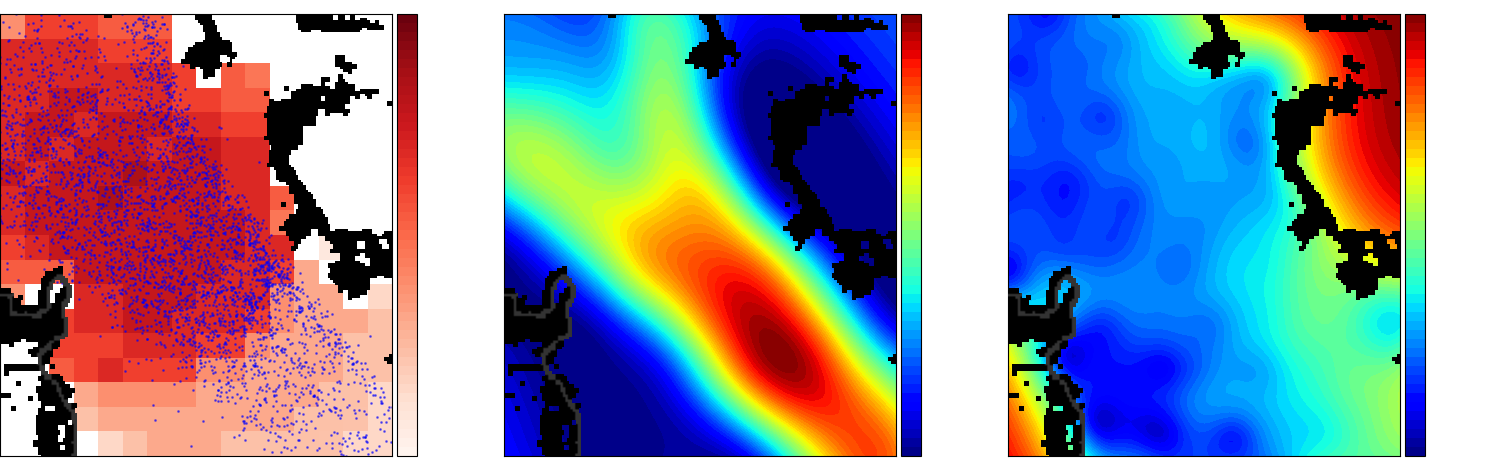}
					};
					
					{\small 
						\setlength{\hgap}{3.0mm}
						%
						\node[header, below right=2.3 mm and 0mm of P.south west, anchor=west] (b1) {(c)}; 
						\node[header, right=\hgap of b1.east, anchor=west] (b2) {(d)}; 
						\node[header, right=\hgap of b2.east, anchor=west] (b3) {(e)}; 

						\newlength{\cbarw}
						\setlength{\cbarw}{2.9cm}
						\newlength{\dbarlab}
						\setlength{\dbarlab}{0.4 mm}
						\colbars{24.2}
					}
					
				\end{tikzpicture}
		\end{center}

		\caption{Upper images: Exemplary 1D dataset with gaps and noise, fitted with a Gaussian process model with Gaussian likelihood (a) and a Gaussian process with a heteroscedastic Gaussian likelihood (b).
		Mobile robots detect a varying number of pedestrians (blue dots), depending on the observation duration at different locations (c).
		Our model aims to infer a continuous rate function of human activity, which compensates for these effects (d).
		Locations with fewer or irregular detections are indicated by the predictive variance (e), based on the heteroscedastic likelihood.}
		\label{fig:intro}
		\vspace{-4.5mm}
	\end{figure}

	The ability to create environmental models is a crucial requirement for the autonomy of mobile robots.
	Especially in long-term applications, the consideration of environmental dynamics has proven to be useful for localization or navigation purposes \cite{Krajnik2017}.
	Human behavior represents a major influencing factor on environmental dynamics, particularly for applications in service robotics or autonomous driving.
	To ensure that robots are accepted by humans and not perceived as a disturbance, they should adapt to human behavior, e.g. in terms of where they move or the timing of their tasks.
	Accurate models of human activity, i.e. spatio-temporal occurrences and movements of pedestrians, can help robots to achieve this purpose, e.g. by improving navigation \cite{OCallaghan2011}, task planning \cite{vintr2020natural} or human-centered task execution \cite{Stuede2021}.
	As current research shows \cite{Vintr2019, vintr2019time}, continuous representations can better reflect human activity than approaches that use spatial or temporal discretization, as interdependencies between data points can be accounted for.
	Following this idea, we present CoPA-Map (\textbf{Co}ntinuous \textbf{P}edestrian \textbf{A}ctivity Map), a non-parametric model for long-term prediction of human presence.
	We focus on the use in mobile robotics, which is characterized by varying dwell times at different locations and thus leads to an inhomogeneous, or sparse, distribution of measurement data.
	The model is implemented using multi-latent \ac{gpr}, allowing time- and location-dependent variances to be incorporated using a heteroscedastic likelihood function.
	Locations with high variability of observed human activity, for example, due to short dwell times, as well as outliers are thereby given lower weight by adjusting the likelihood variance during hyperparameter optimization.
	This can also be accounted for in the resulting predictive uncertainty to indicate areas of the input space which require further exploration or data collection (see Fig.~\ref{fig:intro} for an example).
	\acp{gp} are also particularly suitable for the given application in that spatial correlations or temporal characteristics can be taken into account as prior information.
	Based on a data-driven initialization procedure, we therefore create a multi-dimensional kernel that encodes long-term periodic patterns resulting from people's routines.
	The code of our method is available online \cite{StuedeCopa}.
	
	The remainder of this paper is structured as follows: the next section \ref{sec:state_oa} gives an overview of related work. Sec. \ref{sec:prem} introduces \ac{gpr} and corresponding preliminaries. Sec. \ref{sec:method} presents our method CoPA-Map, which is evaluated in Sec. \ref{sec:exp} and a conclusion is given in Sec. \ref{sec:conc}.
	
	\section{Related work}
	\label{sec:state_oa}
	Approaches to modeling human activity generally consider spatial or temporal variations or a combination of both.
	For an indication of local variability, many models discretize the spatial coordinates, e.g. using a grid, so that different locations are considered separately.
	In \cite{Senanayake2018}  a directional grid map is presented, that probabilistically models long-term human motion through angular directions.
	Angular representations, that also incorporate motion speed and partial observability are presented in \cite{Kucner2017}.
	Instead of separating the environment into discrete locations, other approaches create continuous representations for short-term trajectory predictions as in \cite{Ellis2009} or \cite{Chung2010}.
	In \cite{OCallaghan2011} spatially continuous navigational maps by observation of human trajectories are created, with a particular focus on integrating a prior path enabled by a Gaussian Process framework.
	Apart from the modeling of human activity, spatially continuous models have been successfully used for occupancy mapping of static objects \cite{OCallaghan2012}.
	Later works  \cite{Senanayake2017,senanayake2017bayesian} also incorporate environmental dynamics to create long-term maps of occupancy.
	These non-parametric methods are typically kernel-based and therefore can distinguish well between empty and occupied space, and can also capture nonlinear or obstructed patterns.
	However, the aforementioned works \cite{Senanayake2018, Kucner2017, Ellis2009, Chung2010, OCallaghan2011, OCallaghan2012, Senanayake2017,senanayake2017bayesian} focus on spatial relations and neglect temporal variations, especially with respect to long-term changes.
	Models which consider long-term temporal patterns usually focus on periodic changes, which can be modeled kernel-based \cite{Tompkins2018} or with spectral analysis, e.g. by the FreMEn method \cite{Krajnik2017}.
	FreMEn is a method for non-uniform frequency transforms with an application to mobile robotics and was originally developed to model the evolution of binary states over time, such as cells of an occupancy grid.
	Therefore, extensions have been made to model human activity quantitatively using spatially discrete Poisson processes with respect to intensities \cite{Jovan2016} or predominant directions of human flow \cite{Molina2019}.
	As these methods either only consider temporal variations \cite{Tompkins2018} or neglect interdependencies of separate spatial regions  \cite{Jovan2016, Molina2019}, authors of \cite{Vintr2019} proposed a spatio-temporal continuous model of human presence.
	The model is based on a projection of data points to a circular space with subsequent clustering by \acp{gmm} and was later extended to incorporate human flow \cite{vintr2019time}.
	However, since clustering is performed directly on the data points (people detections), it is prone to erroneous predictions when the robotic system moves through the environment and collects varying amounts of data at different locations. 
	
	In summary, the long-term prediction of human activity, which is suitable for mobile robotic applications, requires further research.
	The contributions of this paper are therefore:
	1) A model for long-term predictions of human presence that compensates for inhomogeneous data distribution resulting from a moving robot and incorporates spatio-temporal interdependencies due to its continuous representation,
	2) a data-specific routine for initializing hyperparameters representing periodic changes in human activity which significantly enhances model convergence,
	3) experiments of the method on real-world datasets.

	
	



	\section{Preliminaries}
	\label{sec:prem}
	\subsection{Gaussian Process Regression (GPR)}
	For a dataset of $n$ training inputs   $\vec{X}=\left\{\vec{x}_i\in \mathbb{R}^d \right\}^n_{i=1}$ and observations $\vec{y}=\left\{y_i\in \mathbb{R} \right\}^n_{i=1}$ the  standard formulation of \ac{gpr} aims at inferring a latent function $f:\mathbb{R}^d \to \mathbb{R}$ via a noisy observation model
	\begin{equation}
		\label{eq:gp_obs_model}
		y_i=f(\vec{x}_i) + \varepsilon_i,\quad \varepsilon_i \sim \mathcal{N}(0, \sigma^2).
	\end{equation}
	The Gaussian Process is defined as a distribution over functions $\rbf{f}=f(\vec{x})\sim\mathcal{GP}(\mu_f(\vec{x}), k_f(\vec{x}, \vec{x}^\prime))$ with mean function $\mu_f(\vec{x})$ and covariance function $k_f(\vec{x}, \vec{x}^\prime)$.

%

	\subsection{Prior Approximation via Inducing Inputs}
	\label{sec:prior_approx}
	The most prominent weakness of standard \acp{gp} is their cubic complexity in the number of training inputs $\mathcal{O}(n^3)$	due to the inversion of the $n \times n$ kernel matrix $\Kff=k_f(\vec{X}, \vec{X})$.
	This limits their usability, especially for applications in robotics and on large datasets.
	A common approach to overcome this problem is to sparsely approximate the kernel matrix  $\Kff$ using the Nyström low-rank representation $\Kff \approx \Kfuf \Kufuf^{-1} \Kfuf^{\transpose}$.
	Therefore, a number of $m$ inducing points (or pseudo-inputs), where $m \ll n$, must be chosen at locations $\rbf{Z}=\left\{\rbf{z}_i\right\}^m_{i=1}$ to optimally represent the training data.
	The corresponding function values are denoted as $\rbf{u}_{f}=f\left(\rbf{Z}\right)$.
	This decreases the computational cost to $\mathcal{O}(m^2n)$, which can further be reduced by variational approximations utilizing \ac{sgd} (see \ref{sec:optimization}).
	As the quality of the approximation largely depends on the number and location of inducing inputs, it is suitable to treat the inducing points as hyperparameters, and optimize their locations $\rbf{Z}$ with respect to the marginal likelihood.
	
	\subsection{Variational Inference for Multiple Latent Functions}
	In the case of heteroscedastic \ac{gpr}, parameters of the likelihood function can vary with the input.
	For Gaussian likelihoods this changes the original \ac{gp} model (eq. \ref{eq:gp_obs_model}) to $y_i \sim \mathcal{N}(f(\vec{x}_i), \zeta\left(g(\vec{x}_i)\right))$, where $g(\vec{x})\sim\mathcal{GP}(\mu_g(\vec{x}), k_g(\vec{x}, \vec{x}^\prime))$ is a second latent function that can also be modeled by a \ac{gp}.
	The function $\zeta\left(.\right):\mathbb{R}^d \to \mathbb{R}^d_{+}$  is a link function to guarantee positive values for the noise parameter \cite{lazaro2011variational}.
	
	In a model with multiple latent functions the marginal likelihood $p(\vec{y})$ is not analytically tractable and posterior approximations are required.
	Instead of calculating the intractable posterior $p(\rbf{f},\rbf{g}\vert\vec{y})$, it can be lower bounded with variational distributions $q(\rbf{f})$ and $q(\rbf{g})$, a technique called \emph{variational inference}.
	The main principle of this technique is the estimation of the parameters of $q(\rbf{f})$ and $q(\rbf{g})$ by minimizing their distance to the true posterior distribution measured by the Kullback-Leibler-divergence $\operatorname{KL}\left(q(\rbf{f})q(\rbf{g}) \| p(\rbf{f},\rbf{g}\vert\vec{y})\right)$.
	Assuming that the latent functions $\rbf{f}$ and $\rbf{g}$ are a priori independent for each data point, Saul et al. \cite{Saul2016} derive the variational lower bound 
	\begin{equation}
		\label{eq:lowerbound}
	\begin{aligned}
		\mathcal{L} = & \sum_{i=1}^{n} \int q\left(\rbf{f}_{i}\right) q\left(\rbf{g}_{i}\right) \log p\left(\vec{y}_{i} \mid \rbf{f}_{i}, \rbf{g}_{i}\right) \diff \rbf{f}_{i} \diff \rbf{g}_{i} \\
		&-\operatorname{KL}\left(q\left(\rbf{u}_{f}\right) \| p\left(\rbf{u}_{f}\right)\right)-\operatorname{KL}\left(q\left(\rbf{u}_{g}\right) \| p\left(\rbf{u}_{g}\right)\right).
	\end{aligned}
\end{equation}
This bound leverages the aforementioned sparse formulation and aims at calculating sparse approximate posteriors as normal distributions $q(\rbf{u}_f)=\mathcal{N}\left(\rbf{u}_{f}\vert \boldsymbol{\mu}_{f}, \boldsymbol{S}_{f}\right)$ and $q(\rbf{u}_g)=\mathcal{N}\left(\rbf{u}_{g}\vert \boldsymbol{\mu}_{g}, \boldsymbol{S}_{g}\right)$ over inducing functions $\rbf{u}_f$ and $\rbf{u}_g$.
For $q(\rbf{f})=\mathcal{N}(\rbf{f}\vert \rbf{m}_f, \rbf{\Sigma}_f)$ follows
\begin{align}
	\rbf{m}_f&=\Kfuf\Kufuf\inv\vec{\mu}_f, \\
	\rbf{\Sigma}_f&=\Kff + \Kfuf\Kufuf\inv(\vec{S}_f-\Kufuf)\Kufuf\inv\Kuff.
\end{align}
The equations for $q(\rbf{g})=\mathcal{N}(\rbf{g}\vert \rbf{m}_g, \rbf{\Sigma}_g)$  follow accordingly.
Training the model is then realized by minimizing $-\mathcal{L}$ with respect to the variational parameters $\vec{\mu}_{f,g}$ and $\vec{S}_{f,g}$ as well as the hyperparameters in the covariance matrices $\Kxx{*}{*}$.
The latter follow from problem-specific covariance functions, which are chosen to account for prior information (see \ref{sec:meth_cov}).

	\section{Methods}
		\label{sec:method}
	To form our model, we consider a mobile robot acting in an environment with a known map,  sufficiently accurate self-localization within this environment and a sensor for people detection.
	A detected pedestrian is represented as a 2D-point $\vec{p}_k=(x_{1, k}, x_{2, k}, t_k)^{\transpose}$ in world coordinates corresponding to a measurement taken at time $t_k$.
	The goal then is to model human activity as an intensity function of space and time,  by first defining a count of people $c_i$ within a spatio-temporal domain $\domst \subset \mathbb{R}^3$ so that $c_i=\lvert\{\vec{p}_k \in \domst\}\lvert$.
	By partitioning the environment into an evenly spaced grid of $n$ cells, we create each domain $\domst$ as a cell with square spatial shape with edge length $r_{\mathrm{s}}$ and temporal resolution $\tau$. 
	Since the robot is moving through the environment, each cell is visible to the robot for a different time period.
	This time period is calculated based on the \ac{fov} of the sensor, which can be approximated by a geometrical shape.
	For example, the projected 2D-detection area of a 3D-Lidar-based detector can be approximated by a circle, which is pruned at known obstacles in the environmental map based on a ray casting model.
	A people count $c_{i} \geq 0$ and observation duration $0 < \Delta_{i}\leq \tau$ is then assigned to each visible cell.
	Consequently, the robot's deployments over time generate the set of input data $\vec{X}=\left\{(x_{1, i}, x_{2, i}, t_i) \right\}_{i=0}^n$, which consists of the spatio-temporal centers of the cells.
	The corresponding target are the observed rates $\vec{y}=\left\{\nicefrac{c_i}{\Delta_i} \right\}_{i=0}^n$ of people in each cell.
	Considering the rates instead of counts is based on the following idea: Since a target value $y_i$ can both be large due to a large $c_i$ or a small $\Delta_i$, it varies more smoothly at edges between areas with shorter and larger observation periods $\Delta_i$.
	As people move through the environment in a continuous fashion, areas with consistent values $y_i$ then indicate homogeneous activity which merits greater weighting when optimizing the marginal likelihood.
	However, irregular spatial patterns of the values in $\vec{y}$ indicate either short observation durations or irregular occurrences of people, which in contrast should be captured by a larger input noise in the likelihood function.
	In Fig.~\ref{fig:overview}, an overview of the input data and resulting rate $\vec{y}$ is given, along with a ground truth which was created without any constraints on the \ac{fov} or observation duration.
	
\subsection{Likelihood Function}
To fit a model in the \ac{gp}-framework, a likelihood function must be chosen that best represents the distribution of observations $\vec{y}$.
Count data, such as person occurrences, can e.g. be viewed as events from an inhomogeneous Poisson process \cite{Jovan2016}.
However, this requires strong assumptions on the independence of events (e.g. people cannot arrive in groups), considers discrete data instead of a continuous rate $y_i$ and the variance of the Poisson distribution is directly coupled to its rate parameter.
Instead, we consider the rate $y_i \sim \mathcal{N}(f(\vec{x}_i), \sigma^2_i)$, to be normally distributed with input-dependent noise $\sigma^2_i$, which can be defined independently from the latent mean function $f(\vec{x}_i)$ and makes training less prone to outliers.
When standardizing the target values $y_i$ to zero-mean and a standard deviation of one, this consistently leads to better results than strictly positive likelihoods, such as the Gamma distribution.
The latter would additionally require manually tuned normalization for different input datasets to achieve consistent results.
As the variance is defined by a latent function $\sigma^2_i= \zeta\left(g(\vec{x}_i)\right)$, we chose the softplus function as link function $\zeta$ to ensure for positive values.
Although the latent function $f(\vec{x}_i)$ can result in negative values, the rescaled predictive output of a tuned model contained very few zero-crossings on all tested datasets, making it sufficient to use the absolute value of the model output for predictions.

\subsection{Definition of Covariance Functions}
\label{sec:meth_cov}
Covariance functions allow encoding prior beliefs about the latent function of interest and can be viewed as a measure of how \emph{similar} two functions are.
Different suitable covariance functions can also be connected as compositions.
For the present use case of representing human activity, each data point is separated into its spatial component $\xs\in\mathbb{R}^2$ and temporal component $\xt\in\mathbb{R}$ and the following covariance function is defined
	\begin{equation}
		k_f\left(\xs, \xt, \xs^{\prime}, \xt^{\prime}\right)=k_{\mathrm{s}}\left(\|\xs-\xs^\prime\|_2\right)k_{\mathrm{t}}\left(\lvert\xt-\xt^\prime\lvert\right).
	\end{equation}
This multidimensional product kernel connects a spatial component $k_{\mathrm{s}}$ with a temporal component $k_{\mathrm{t}}$ and results in a prior over functions that varies across all three dimensions.
As the spatial kernel, the Matérn-$\nicefrac{5}{2}$ covariance function
\begin{equation}
	k_{\mathrm{s}}(r)=\sigmas^2\left(1+\frac{\sqrt{5} r}{\ls}+\frac{5 r^{2}}{3 \ls^{2}}\right) \exp \left(-\frac{\sqrt{5} r}{\ls}\right)
\end{equation}
is chosen, where $\ls$ and $\sigmas^2$ are hyperparameters.
This type of covariance function is a common choice to model structural correlations, as it provides a good balance between smoothness and capturing sudden changes \cite{kim2013continuous}.

Oftentimes, human activity can be considered periodic in time.
The number of people at different locations is subject to a regularity that is determined, for example, by the time of day, working hours or store opening hours.
Therefore, as prior information for the time-dependent person rate, we specify the rate to be subject to periodicities.
This can be encoded by a periodic kernel \cite{mackay1998introduction}, which is defined as a sum of trigonometric functions
\begin{equation}
	\label{eq:kt}
	k_{\mathrm{t}}\left(r\right)=\sum_{i=0}^{\psi}\sigma_{\mathrm{t},i}^{2} \exp \left(-\frac{1}{2} \frac{\sin^2 \left(\gamma_i\inv r\right)}{l_{\mathrm{t},i}^2}\right)
\end{equation}
where the variances $\sigma_{\mathrm{t},i}^{2}$, periods $\gamma_i$ and lengthscales $l_{\mathrm{t},i}$ are hyperparameters.
The variances $\sigma_{\mathrm{t},i}^{2}$ determine the overall influence of the specific component and $l_{\mathrm{t},i}$ controls the smoothness.

Regarding human activity, the kernel $k_f$ thus represents two important properties: 1.) Spatial continuity, i.e. if people are seen at a specific location it is more likely to also see people at locations that are very close.
Since humans move through space in a continuous manner, this property is desirable to model.
2.) Temporal periodicity, i.e. when people are seen repeatedly at a specific location (e.g. every morning at an entrance) it is likely to see people there in the future at that specific point in time.
The kernel $k_g$ corresponding to the latent function $\rbf{g}$ is simply realized by a \ac{rbf} kernel.
This is sufficient since then the predictive variances of different areas align for larger prediction horizons.

\subsection{Initialization of Hyperparameters}
Due to the dependence on many data points as well as hyperparameters, optimization of the lower bound (eq. \ref{eq:lowerbound}) is prone to get stuck in local minima. 
A major influencing factor is the initial guess of the hyperparameters.
In the present scenario, this applies in particular to the periods $\gamma_i$ and variances $\sigma_{\mathrm{t},i}^{2}$ of the temporal kernel $k_{\mathrm{t}}$ (eq. \ref{eq:kt}).
With algorithm \ref{alg:per}, we therefore propose a method to obtain the characteristic temporal periods of a spatial domain based on non-uniform frequency analysis and a subsequent clustering step.
The algorithm builds on the idea  \cite{Krajnik2017} of transferring the time-dependent activities at different locations into the frequency spectrum and making an approximation via a Fourier series with a reduced number of components.
By squashing the cells $\domst$ of the spatio-temporal grid along the temporal dimension, a spatial grid with a time series of rates $\vec{y}_s \subset \vec{y}$ for each spatial cell $s$ results.
A subset $\doms$, containing a number of $l$ spatial cells, is then taken from this spatial grid by sampling, where each cell is given a weight of its total counts over all timesteps.
This results in the selection of cells that are more likely to have high activity but does not completely exclude cells with lower activity.
Due to the movement of the robot, the rates within $\vec{y}_s$ are non-equidistant with respect to the time of their detection.
The conversion to the frequency domain is therefore made by means of the \ac{nudft} \cite{dutt1993fast} (line \ref{per:nudft}).
This requires a set of candidate periods $O$, which is defined to contain equally spaced periods within an interval (e.g. between one hour and seven days).
Additionally, the algorithm needs an upper limit $\pmax$ of periods to check and a scaling factor $\sigma^2_{\mathrm{max}}$ as the maximum variance.
The optimal number of periods for each cell is determined by five-fold cross-validation, by comparing the test data with the signal that was reconstructed from a reduced number of frequency components (lines \ref{per:cv} to \ref{per:endcell}).
The total number of periods $\psi$ of the whole domain is then calculated as the mean of the number of periods of the cells in $\doms$ (line \ref{per:wm}).
The periods are calculated by weighted $k$-means clustering, where the complex magnitudes serve as weights (line \ref{per:kmeans}).
This ensures that locations with a large recurring number of people are more influential than cells that have less activity.


\begin{figure}[t]

	\removelatexerror
		\vspace{2mm}
	\begin{algorithm}[H]
	\newcommand\mycommfont[1]{\footnotesize\ttfamily{#1}}
	\SetCommentSty{mycommfont}
	\caption{Init. hyperparameters of periodic kernel}\label{alg:per}
	{\small 
		\SetKwInOut{Input}{Input}
		\SetKwInOut{Output}{Output}
		\Input{$\vec{y}$, $O$, $\doms$, $\pmax$, $\sigma^2_{\mathrm{max}}$}
		\Output{$\psi$, $\hat{\gamma}_{1..\psi}$, $\hat{\sigma}^2_{1..\psi}$}
		\ForEach{$s \in \doms$}{
			Let $\vec{y}_s$ be the rates at times $\vec{t}_s$ of spatial cell $s$\;
			Repeat lines \ref{per:split} -- \ref{per:endcell} as  cross validation for $i=1..5$\label{per:cv}\;
			Split $\vec{y}_s$ into contiguous train/test sets $\vec{y}_s^{\mathrm{tr}}/\vec{y}_s^{\mathrm{ts}}$\label{per:split}\;
			$\vec{\xi}_i \gets \mathrm{NUDFT}(\vec{t}_s^{\mathrm{tr}}, \vec{y}_s^{\mathrm{tr}}, O)$\label{per:nudft}\tcp*{To cplx. components}
			\For{$p=0$ $\mathrm{to}$ $\pmax$}{
				$\vec{\xi}_{i,p} \gets$ $p$ largest complex numbers in $\vec{\xi}_i$ w.r.t. magnitude\;
				$O_{i,p} \gets$ Set of periods, corresponding to $\vec{\xi}_{i,p}$\;
				$\hat{\vec{y}}_p \gets \mathrm{InverseDFT}(\vec{\xi}_{i,p}, O_{i,p})$\;
				$e_{i,p} \gets \mathrm{RMSE}(\hat{\vec{y}}_p, \vec{y}_s^{\mathrm{ts}})$\;
			}\label{per:endcell}
			
			$i_s, p_s \gets \argmin_{i,p}(e_{1,0},..., e_{5,\pmax})$ \;
			$A_s \gets $ Save element-wise magnitudes of  $\vec{\xi}_{i_s, p_s}$\;
			$O_s \gets $ Set of periods, corresponding to $\vec{\xi}_{i_s, p_s}$\;
		}
		
		$\psi \gets \lfloor\mathrm{Mean}(\left\{p_1, ..., p_{\lvert\doms\lvert} \right\})\rfloor$\label{per:wm}\;
		$\hat{\gamma}_{1..\psi} \gets $ obtain $k$-$\mathrm{means}$ centroids with $k=\psi$ using $\left\{O_s \mid s \in \doms \right\}$ with weights  $A_s$\label{per:kmeans}\;
		$\hat{\sigma}^2_{1...\psi} \gets$ Sum weights $A_s$ in clusters and normalize to $\left[0,\sigma^2_{\mathrm{max}}\right]$\label{per:sum_cl}\;
	}
	
  \end{algorithm}
		\vspace{-6mm}
\end{figure}
\begin{figure*}[t]
	
	\noindent
	\newcommand{\colarr}[1]{
		\node[inner sep=0pt,#1, anchor=south] (lower) {};
		\node[scale=1, above = 11mm of lower] (upper) {};
		\draw[line width=1pt, -{Latex[length=2mm,width=1.4mm]}] (lower.south) -- (upper.south);
	}
	\newcommand{\colbars}[1]{
		\colarr{below right= #1mm and 0.0mm of h1.south east}
		\node[inner sep=0pt, align=center, right= \dbarlab of lower.south, anchor=south west] (c1)  {$\Delta_i$}; 
		\colarr{right= \cbarw of lower.south}
		\node[inner sep=0pt, align=center, right= \dbarlab of lower.south, anchor=south west] (c2) {$y_i$}; 
		\colarr{right= \cbarw of lower.south}
		\node[inner sep=0pt, align=center, right= \dbarlab of lower.south, anchor=south west] (c3)  {$y_{\mathrm{gt}}$}; 
		\colarr{right= \cbarw of lower.south}
		\node[inner sep=0pt, align=center, right= \dbarlab of lower.south, anchor=south west] (c4)  {$\rbf{m}_y$}; 
		\colarr{right= \cbarw of lower.south}
		\node[inner sep=0pt, align=center, right= \dbarlab of lower.south, anchor=south west] (c5)  {$\vec{\sigma}_y$}; 
	}
	\begin{center}
		\flushleft
		\vspace{-3.3mm}
		\resizebox{1.00\textwidth}{!}{
			\begin{tikzpicture}
				\tikzset{
					header/.style={
						text width=.158\textwidth,
						align=center,
						inner sep=0pt,
						anchor=west
					}
				}
				\node[inner sep=0pt, anchor=south west, align=left] (P) at (-2,0)
				{\includegraphics[trim=0 3 0 3, clip, width=0.93\textwidth]{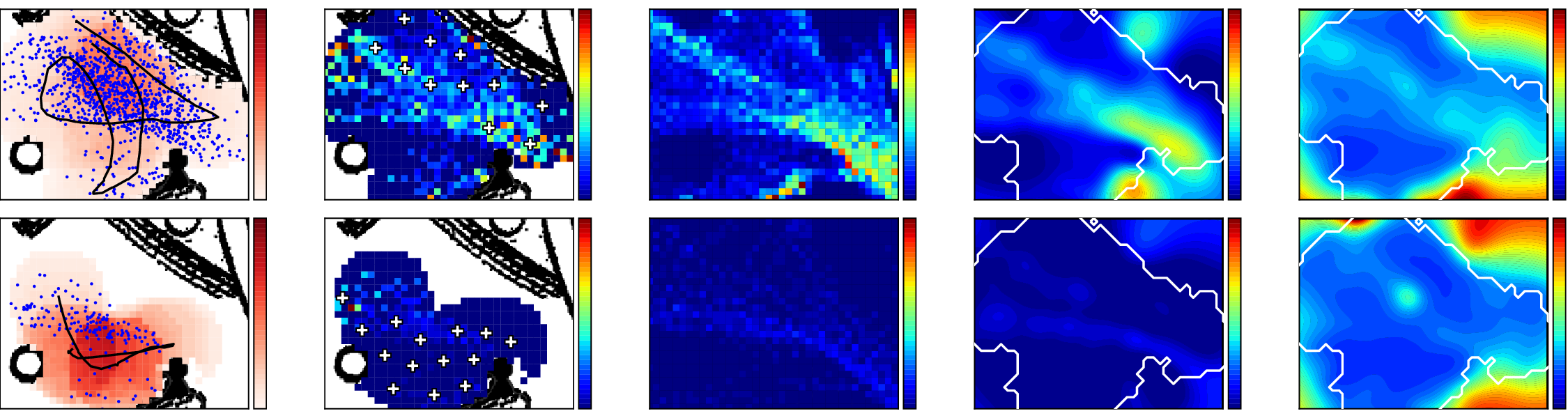}
				};
				
				{\small 
					\node[above left=23.1 mm and 4mm of P.south west, anchor=west,rotate=90, text width=1.86cm, align=center] (t1) {1:00 p.m. to 2:00 p.m.}; 
					\node[above left=0.0 mm and 4mm of P.south west, anchor=west,rotate=90, text width=1.86cm, align=center] (t2) {8:00 p.m. to 9:00 p.m.}; 
					\setlength{\hgap}{6.0mm}
					\node[header, above right=3.5 mm and 0mm of P.north west, anchor=west] (h1) {Robot path and detected people}; 
					\node[header, right=\hgap of h1.east, anchor=west] (h2) {Observed rates $\vec{y}$ and inducing points $\hat{\rbf{Z}}$}; 
					\node[header, right=\hgap of h2.east, anchor=west] (h3) {Ground truth}; 
					\node[header, right=\hgap of h3.east, anchor=west] (h4) {Model output: Mean}; 
					\node[header, right=\hgap of h4.east, anchor=west] (h5) {Model output: Std. deviation}; 
					
					\node[header, below right=2.5 mm and 0mm of P.south west, anchor=west] (b1) {(a)}; 
					\node[header, right=\hgap of b1.east, anchor=west] (b2) {(b)}; 
					\node[header, right=\hgap of b2.east, anchor=west] (b3) {(c)}; 
					\node[header, right=\hgap of b3.east, anchor=west] (b4) {(d)}; 
					\node[header, right=\hgap of b4.east, anchor=west] (b5) {(e)}; 
				}
				
				\setlength{\cbarw}{3.424cm}
				\setlength{\dbarlab}{0.3 mm}
				
				\colbars{20.9}
				\colbars{42.8}
				\begin{pgfonlayer}{bg}
					\node[below left =1.5mm and 10mm of t1.south west, fill=black!30!white, align=left, draw=black, anchor=south west,minimum height=2.18cm,minimum width=18.0cm] (bg1) {};
					\node[fit={(bg1)}, below =0mm of bg1.south west, fill=black!10!white, align=left, draw=black, anchor=north west, inner sep=0pt] (bg2) {};
				\end{pgfonlayer}
			\end{tikzpicture}
		}
	\end{center}

	\caption{Input and output data (ATC dataset) of the model at midday (top) and evening (bottom). People detections $\vec{p}_k$ (blue) and observation durations $\Delta_i$ are shown in (a). Resulting rates $\vec{y}$ for a bin duration of $\tau=60$\,min and initial inducing points (white plusses) are shown in (b). A ground truth with fully observed cells is given for reference in (c). The resulting model outputs are shown in (d) and (e) with a white border indicating the areas that were never observed. }
	\label{fig:overview}
	\vspace{-3mm}
\end{figure*}

As the full covariance matrix $\Kff$ is not computed, but approximated by covariances over inducing points, their positioning is an additional factor influencing the model quality.
Although the inducing points are treated as hyperparameters and therefore modified during optimization, proper initialization reduces the time to find sufficient solutions.
Given a ratio $\alpha\in(0, 1]$,  the number of inducing points is selected as $m=\lfloor\alpha n\rfloor$.
The location is then determined via $k$-means clustering ($k=m$) of the spatio-temporal training inputs $\vec{X}$, where each input point is weighted by its individual observation time $\Delta_i$.
By weighting the inputs, the initial inducing points $\hat{\rbf{Z}}$ resulting from the algorithm are primarily placed at locations that have been observed for longer periods and hence provide more reliable data.
An exemplary arrangement of inducing points is shown in Fig.~\ref{fig:overview}~(b).

\subsection{Model Optimization}
\label{sec:optimization}
As indicated in section \ref{sec:prior_approx}, the optimization of the hyperparameters does not scale cubically when latent inducing locations $\rbf{Z}$ are used.
When the covariance matrices of the variational distributions are parametrized as Cholesky $\vec{S}_f=L_{\rbf{f}}L_{\rbf{f}}^\transpose$ and $\vec{S}_g=L_{\rbf{g}}L_{\rbf{g}}^\transpose$, optimizing the lower bound $\mathcal{L}$ (eq. \ref{eq:lowerbound}) scales with $\mathcal{O}(nm^2+2nm)$ \cite{Saul2016}.
By choosing a ratio parameter of $\alpha$ so that $m \ll n$, model inference is significantly sped up compared to the standard \ac{gpr} case.
We empirically chose a parameter of $\alpha=0.02$ to obtain a good balance between computational speed and prediction quality on the evaluated datasets.
Model optimization is executed for three types of parameters: 1.) Variational parameters corresponding to $q(\rbf{u}_f)=\mathcal{N}\left(\rbf{u}_{f}\vert \boldsymbol{\mu}_{f}, \boldsymbol{S}_{f}\right)$ and $q(\rbf{u}_g)=\mathcal{N}\left(\rbf{u}_{g}\vert \boldsymbol{\mu}_{g}, \boldsymbol{S}_{g}\right)$, 2.) the lengthscale, variance and periodicity hyperparameters of the covariance functions $k_f$ and $k_g$, and 3.) the location of inducing inputs $\rbf{Z}$.
For this, two separate optimization techniques based on \ac{sgd} are utilized.
The variational parameters are optimized using the natural gradient method since the inherent minimization of KL-divergence as part of this method integrates well with the variational framework and leads to fast convergence \cite{salimbeni2018natural}.
The kernel hyperparameters and inducing point locations are optimized with the Adam optimizer \cite{kingma2014adam}.
Steps of both optimizers are executed in an alternating fashion, with a linearly decaying learning rate for the first 100 optimization steps.
The $n$ integrals as part of the lower bound $\mathcal{L}$ are solved by two-dimensional Gaussian quadratures.
As the methods utilize \ac{sgd}, the optimization can efficiently be separated into mini-batches.  

\subsection{Predicting with the Model}
After maximization of the variational lower bound and optimization of hyperparameters, the model can be queried via its predictive distribution.
For arbitrary new data inputs $\vec{X}^*=\left\{\vec{x}_i^*\right\}_{i=1}^{n^*}$ the predictive distribution is given as $\int p\left(\vec{y}_{i}^{*} \mid \rbf{f}_{i}^{*}, \rbf{g}_{i}^{*}\right) q\left(\rbf{f}_{i}^{*}\right) q\left(\rbf{g}_{i}^{*}\right) \diff \rbf{f}_{i}^{*} \diff \rbf{g}_{i}^{*}$.
This analytically intractable integral can be computed using Gauss-Hermite quadrature to obtain the predictive mean $\rbf{m}_y$ and variance $\vec{\sigma}^2_y$.
The specific values of the predictive mean depend on the chosen spatial resolution $r_\mathrm{s}$ and temporal resolution $\tau$ of the input grid.
For a subset $\vec{X}^\prime \subset \vec{X}^*$ of finite extend (e.g. the \ac{fov} of the robot and a given duration) the expected number of people can then be calculated as a point estimate $\frac{1}{r_{\mathrm{s}}^2\tau}\int\rbf{m}_y \diff\vec{X}^\prime$.
Exemplary model outputs of the mean $\rbf{m}_y$ and standard deviation $\vec{\sigma}_y$ for two points in time are shown in Fig.~\ref{fig:overview}~(d) and (e).
The uncertainty increases both outside the visited area and in locations where high variability of human activity occurs.
In addition to the predictive uncertainty, this indicates in which areas further model exploration could be useful.
	\section{Experiments}

		\label{sec:exp}
All the following experiments were performed with the same parameterization: $l=10$, $\psi_{\mathrm{max}}=10$, $\sigma^2_{\mathrm{max}}=0.95$, $\alpha=0.02$.
The initialization routine (Algorithm \ref{alg:per}) was done with a fixed grid resolution of 5.0\,m $\times$ 60\,min, whereas the grid resolution resulting in $\vec{X}$ was varied for different experiments (respectively specified).
The method is implemented in Python based on the GPflow library \cite{Matthews2017} to perform the training and inference GPU-based.

\subsection{Datasets}
We evaluated the model on two freely available long-term datasets containing real-world pedestrian detections.
Both datasets represent typical settings for mobile robots but vary in terms of human activity and the number of pedestrians.

\quad\emph{ATC Dataset} \cite{brvsvcic2013person}: This dataset contains measurements of tracked pedestrians in a shopping center in Osaka, Japan, covering an area of ca. 900\,$\mathrm{m}^2$.
Data collection was done with multiple 3D range sensors, every week on Wednesdays and Sundays, resulting in 92 days in total.
We downsampled the data to a detection rate of \SI{0.5}{\hertz}, resulting in an average of about 1700 entries per square meter and day.
For evaluation, we used a subset of 10 Wednesdays for training and 4 days for testing.

\quad\emph{Office Dataset} \cite{molina2021robotic}:  The second dataset contains tracks of people based on measurements by a single stationary 3D-Lidar in an office environment of the University of Lincoln, England, covering an area of ca. 85\,$\mathrm{m}^2$ with averagely about 300 entries per square meter and day.
The dataset covers 22 consecutive days, of which we used 10 weekdays for training and 5 weekdays for testing.

As both datasets contain measurements taken by stationary sensors, data collection by a moving robotic system must be simulated.
For this purpose, robot trajectories with an average moving speed of $\SI{0.5}{\meter\per\second}$ and intermediate stationary stops were specified manually.
Then, only the measurements within the \ac{fov} of the robot are processed.
The \ac{fov} is defined by a circle with a fixed radius and is pruned based on the known occupancy maps of the environments to filter out pedestrians that would be obstructed by static obstacles.
Exemplary sets of measurements and robotic paths are shown in Fig.~\ref{fig:overview}~(a).

\subsection{Evaluation Metrics and Baselines}

The predictive quality of the model is measured with three criteria.
As the evaluation is conducted based on multiple paths, each with different length and area coverage, the first criterion is \ac{nrmse} between model predictions $\hat{y}_i$ and ground truth $y_{\mathrm{gt},i}$ 
\begin{equation}
	\mathrm{NRMSE}=\sqrt{\frac{1}{\bar{y}_{\mathrm{gt}}^2\,n_{\mathrm{test}}}\sum_{i=1}^{n_{\mathrm{test}}}\left(\hat{y}_i-y_{\mathrm{gt},i}\right)^2},
\end{equation}
normalized by the mean test data value $\bar{y}_{\mathrm{gt}}$.
The ground truth value is obtained in a similar way as the creation of training data $\vec{y}$, but ignoring the occupancy map and setting the observation durations $\Delta_i=\tau$.
It therefore represents  the people count $c_{\mathrm{gt},i}$ during testing time in the cells of the spatio-temporal grid that were visited during training, using the same spatial and temporal resolution.
The second criterion is the Chi-square distance
\begin{equation}
	\chi^2\mathrm{-distance}=\sum_{i=1}^{n_{\mathrm{test}}}\frac{\left(\hat{y}_i-y_{\mathrm{gt},i}\right)^2}{\left(\hat{y}_i+y_{\mathrm{gt},i}\right)},
\end{equation}
where larger values indicate less accurate prediction compared with the test data.
These two metrics can be regarded as the standard metrics when comparing human activity or flow models \cite{Jovan2016,molina2021robotic,vintr2019time, vintr2020natural}.
However, these criteria have limited expressiveness in terms of the \emph{usefulness} of the model e.g. for supporting unobstructed navigation or task planning.
Vintr et al. \cite{vintr2020natural} therefore proposed new criteria to evaluate these models based on their ability to support human-aware navigation.
The benchmark's main idea is to rate models better which avoid disturbance of people by executing movements of the robot outside of their immediate walking paths.
The criterion considers a number of $p$ imaginary navigation scenarios, where a robot should navigate between a set of goal locations at different points in time.
The navigational path is planned based on the output of the respective model, where higher activity corresponds to higher path costs.
All resulting paths are then ordered ascending by their total cost and the \emph{service disturbance}
\begin{equation}
E(\lfloor pr \rfloor)=\sum_{k=1}^{\lfloor pr \rfloor}e_k
\end{equation}
is defined as a sum of robot-human encounters $e_k$ during test time.
Robot-human encounters $e_k$ are the person detections that occur within a \SI{1}{\meter} radius to the robot while it is simulatively traveling the path at a speed of \SI{0.5}{\meter\per\second}.
The value $r\in\left[0,1\right]$ is referred to as \emph{servicing ratio} and defines the number of navigation actions that should be performed.
A lower servicing ratio gives more freedom to the robot to discard paths that have high costs, e.g. when the number of expected people is large.

Besides CoPA-Map, the following methods are compared in the evaluation.

The \emph{Maximum-Likelihood} (ML) model calculates the mean of all observed rates in each cell. As a result, the rates are assumed to be constant over time.

\emph{Poisson spectral model} \cite{Jovan2016} (Fr-AAM) is a state-of-the-art approach, modeling human activity as an inhomogeneous Poisson process by a spatial grid with a temporally continuous rate function.
For each cell of the grid, a spectral analysis based on the FreMEn method \cite{Krajnik2017} is performed repeatedly to obtain the most influential spectral components, from which the predictive signal is then reconstructed.

\emph{Warped-Hypertime} \cite{vintr2019time} (WHyTe) is a state-of-the-art approach for continuous activity and flow modeling.
It is based on a frequency analysis by the FreMEn method and subsequent projection into a circular space.
As training data, it directly uses people detections $\vec{p}_k$ and outputs the probability of occurrence given an input point.
Because of this, we do not directly compare the model output to the quantitative value  $c_{\mathrm{gt},i}$, but only include this method in the evaluation of service disturbance.
As the method uses a pre-defined number of clusters, we separately trained models with up to seven clusters and only include the variant with the best result.
The method is also capable of estimating movement direction and speed, although this is not used in the present evaluation to ensure direct comparability to the other models.

The \emph{Gaussian Process model} (GP-Hom) is based on the same parameters as the proposed method but realized as a Log Gaussian Cox process.
The method uses a homoscedastic Poisson likelihood for inference by using a single latent function, transformed with an exponential function to only output positive values which is required for the rate parameter of the Poisson distribution.

\subsection{Validating Hyperparameter Initialization}
\begin{figure}[t]
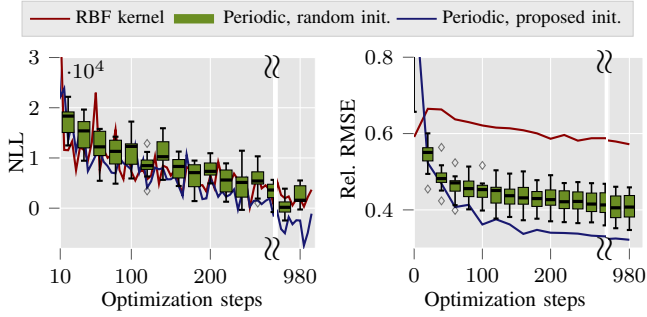

	\vspace{2mm}
	\noindent\makebox[\linewidth][c]{
		\begin{minipage}[t]{0.99\linewidth}
			\centering
			
			\resizebox{0.99\linewidth}{!}{
				
				\centering
				\begin{tikzpicture} 
	\definecolor{color0}{rgb}{0.0980392156862745,0.0980392156862745,0.43921568627451}
	\definecolor{color1}{rgb}{0.419607843137255,0.556862745098039,0.137254901960784}
	\begin{axis}[%
		hide axis,
		xmin=10,
		xmax=12,
		ymin=0,
		ymax=0.4,
		legend cell align={left},
		legend columns=3,
		legend pos=north east,
		legend style={
			fill opacity=0.8,
			draw opacity=1,
			text opacity=1,
			at={(0,10)},
			draw=white!80!black,
			fill=white!89.8039215686275!black
		},
		legend style={font=\footnotesize},
		]
		\addlegendimage{thick, red!54.5098039215686!black}
	\addlegendentry{RBF kernel}
	\addlegendimage{line width=4pt, color1}
		\addlegendentry{Periodic, random init.}
	\addlegendimage{thick, color0}
	\addlegendentry{Periodic, proposed init.}

	\end{axis}
\hspace{-1mm}
\end{tikzpicture}
			}
			
	\end{minipage}}
	\begin{subfigure}[b]{0.49\linewidth}
		\centering
		\setlength{\figureheight}{0.6\linewidth}
		\setlength{\swidth}{0.67\linewidth}
		\centering
		\input{fig/period_init_loss.tex}
		\vspace{-2mm}
	\end{subfigure}
	\hfill
	\begin{subfigure}[b]{0.49\linewidth}
		\centering
		\setlength{\figureheight}{0.6\linewidth}
		\setlength{\swidth}{0.6\linewidth}
		\centering
		\input{fig/period_init.tex}
		
		\vspace{-2mm}
	\end{subfigure}
	
	\caption{NLL and relative RMSE (lower is better) for an RBF kernel, a periodic kernel with the proposed initialization routine and periodic kernels with randomly initialized parameters. Both the ML and CoPA-Map model were trained with a 0.5\,m $\times$ 60\,min resolution.}
	\label{fig:period_init}
				   \vspace{-4mm}
\end{figure}

The first experiment demonstrates the importance of proper initialization of the hyperparameters of the temporal kernel $k_{\mathrm{t}}$.
Given the data from an \SI{185}{\meter\squared} area of the ATC dataset, the model was trained with different periodic kernels and an \ac{rbf}-kernel for comparison.
Besides our proposed initialization procedure (Alg. \ref{alg:per}), we used ten different periodic kernels with variances chosen uniformly randomly in $(0,1)$ and one to two random periods as multiples of 30 minutes and smaller than 30 hours.
Our initialization procedure results in two periods of 12 and 6 hours with variances of 0.9 and 0.42 respectively.
Figure \ref{fig:period_init} shows the \ac{nll} loss during training and the RMSE relative to the ML model on the four independent test days.
Due to the high variance of the training data, NLL shows little variation for the periodicity parameter.
However, suitable parameters of the periodicities lead to significantly better extrapolations, which is reflected in the values of the RMSE.
A complete disregard of periodicities (\ac{rbf} kernel) further results in unsatisfactory predictive results, since long-term changes cannot be captured and the predictive horizon is limited by the kernel's lengthscale parameter.
\begin{table*}[b]
		\vspace{-3mm}
	\caption{Predictive performance of the evaluated models for a static and moving robot. \ac{nrmse} is given as mean and $\chi^2$-distance as a sum over the results from different paths/locations. $\chi^2$-distance is given as multipliers of $10^4$ for brevity of notation.}
	\label{tab:paths_results}
	\resizebox{\textwidth}{!}{
		{\small
			\begin{tabular}{ll>{\raggedleft}p{0.6cm}>{\raggedleft}p{0.75cm}>{\raggedleft}p{0.6cm}>{\raggedleft}p{0.75cm}>{\raggedleft}p{0.6cm}>{\raggedleft}p{0.75cm}>{\raggedleft}p{0.6cm}>{\raggedleft}p{0.75cm}|>{\raggedleft}p{0.6cm}>{\raggedleft}p{0.75cm}>{\raggedleft}p{0.6cm}>{\raggedleft}p{0.75cm}>{\raggedleft}p{0.6cm}>{\raggedleft}p{0.75cm}>{\raggedleft}p{0.6cm}>{\raggedleft\arraybackslash}p{0.75cm}}
\toprule
                                 &          & \multicolumn{8}{c|}{Office} & \multicolumn{8}{c}{ATC} \\
                                 &          & \multicolumn{2}{c|}{{\scriptsize{\centering\arraybackslash}\parbox{1.7cm}{0.5\,m$\times$30\,min}}} & \multicolumn{2}{c|}{{\scriptsize{\centering\arraybackslash}\parbox{1.7cm}{0.5\,m$\times$60\,min}}} & \multicolumn{2}{c|}{{\scriptsize{\centering\arraybackslash}\parbox{1.7cm}{0.75\,m$\times$30\,min}}} & \multicolumn{2}{c|}{{\scriptsize{\centering\arraybackslash}\parbox{1.7cm}{0.75\,m$\times$60\,min}}} & \multicolumn{2}{c|}{{\scriptsize{\centering\arraybackslash}\parbox{1.7cm}{0.5\,m$\times$30\,min}}} & \multicolumn{2}{c|}{{\scriptsize{\centering\arraybackslash}\parbox{1.7cm}{0.5\,m$\times$60\,min}}} & \multicolumn{2}{c|}{{\scriptsize{\centering\arraybackslash}\parbox{1.7cm}{0.75\,m$\times$30\,min}}} & \multicolumn{2}{c}{{\scriptsize{\centering\arraybackslash}\parbox{1.7cm}{0.75\,m$\times$60\,min}}} \\
                                 &          &                                                      {\scriptsize NRMSE} & {\scriptsize $\chi^2\mathrm{dst}$} &                                                      {\scriptsize NRMSE} & {\scriptsize $\chi^2\mathrm{dst}$} &                                                       {\scriptsize NRMSE} & {\scriptsize $\chi^2\mathrm{dst}$} &                                                       {\scriptsize NRMSE} & {\scriptsize $\chi^2\mathrm{dst}$} &                                                      {\scriptsize NRMSE} & {\scriptsize $\chi^2\mathrm{dst}$} &                                                      {\scriptsize NRMSE} & {\scriptsize $\chi^2\mathrm{dst}$} &                                                       {\scriptsize NRMSE} & {\scriptsize $\chi^2\mathrm{dst}$} &                                                       {\scriptsize NRMSE} & {\scriptsize $\chi^2\mathrm{dst}$} \\
\midrule
\multirow{4}{*}{\rotatebox[origin=c]{90}{Static}} & ML &                                               2.66 &                   8.80 &                                               2.34 &                   8.01 &                                      \textbf{2.63} &                   8.66 &                                      \textbf{2.25} &                   7.73 &                                               1.71 &                 415.33 &                                               1.76 &        \textbf{504.95} &                                               1.73 &                 515.34 &                                                1.7 &                 517.45 \\
                                 & Fr-AAM &                                               2.71 &                   8.57 &                                                2.4 &                   7.82 &                                               2.76 &                   9.12 &                                               2.33 &                   8.07 &                                      \textbf{1.71} &        \textbf{412.19} &                                                1.9 &                 506.18 &                                               1.75 &        \textbf{512.83} &                                               1.79 &        \textbf{516.97} \\
                                 & GP-Hom &                                               2.75 &                   8.05 &                                               2.48 &                   8.88 &                                               2.79 &                   8.69 &                                               2.42 &                   9.02 &                                               2.17 &                 528.62 &                                                1.7 &                 524.94 &                                      \textbf{1.68} &                 516.96 &                                      \textbf{1.66} &                 525.44 \\
                                 & CoPA-Map &                                      \textbf{2.59} &          \textbf{6.74} &                                      \textbf{2.33} &          \textbf{6.50} &                                               2.65 &          \textbf{6.95} &                                               2.32 &          \textbf{6.94} &                                               1.78 &                 524.29 &                                      \textbf{1.69} &                 578.23 &                                                1.8 &                 655.80 &                                                1.7 &                 603.43 \\
\cline{1-18}
\multirow{4}{*}{\rotatebox[origin=c]{90}{Moving}} & ML &                                               2.61 &                  35.27 &                                               2.34 &                  32.03 &                                               2.57 &                  35.75 &                                               2.77 &                  38.89 &                                               1.79 &                 456.92 &                                               1.94 &                 460.32 &                                               1.84 &                 477.26 &                                                2.0 &                 481.37 \\
                                 & Fr-AAM &                                                2.6 &                  33.72 &                                               2.38 &                  30.76 &                                               2.61 &                  33.89 &                                               2.67 &                  34.89 &                                      \textbf{1.36} &        \textbf{345.00} &                                               1.31 &                 438.58 &                                               1.32 &                 420.39 &                                               1.28 &                 429.97 \\
                                 & GP-Hom &                                      \textbf{2.58} &         \textbf{31.35} &                                               2.34 &                  33.49 &                                               2.56 &                  33.60 &                                               2.32 &                  34.74 &                                               1.57 &                 649.01 &                                                1.5 &                 639.74 &                                               1.41 &                 567.46 &                                               1.37 &                 576.19 \\
                                 & CoPA-Map &                                               2.61 &                  50.52 &                                      \textbf{2.13} &         \textbf{24.91} &                                      \textbf{2.33} &         \textbf{25.28} &                                      \textbf{2.19} &         \textbf{25.20} &                                               1.44 &                 606.58 &                                      \textbf{0.82} &        \textbf{172.07} &                                       \textbf{0.8} &        \textbf{164.56} &                                      \textbf{0.69} &        \textbf{127.15} \\
\bottomrule
\end{tabular}

		}
	}

\end{table*}
\subsection{Spatio-Temporal Prediction}
In order to evaluate the predictive quality of CoPA-Map, we consider the two different scenarios of a \emph{static} and \emph{moving} robot.
In the first, a permanently motionless robot is assumed, and a total of five different positions of the robot are considered separately leading to a constant observation time of $\Delta_i = \tau$ for every cell.
For the \emph{moving} case, we created 9 different paths with varying spatial coverage of the robot's \ac{fov} (between \SI{40}{\meter\squared}--\SI{70}{\meter\squared} for Office and \SI{100}{\meter\squared}--\SI{200}{\meter\squared} for ATC datasets) and different waiting times along the paths.
Thus, different cases are covered, where some locations may be permanently in the robot's \ac{fov} and others may be visited only a few times a day.
Table \ref{tab:paths_results} shows the results for \ac{nrmse} and $\chi^2$-distance.
Since these metrics depend on the chosen spatial and temporal resolution ($r_{\mathrm{s}}$ and $\tau$), four different combinations are shown.
For the \emph{static} case of the Office dataset, CoPA-Map generally leads to better results than the comparative methods.
In the ATC experiments, there is a location in the static case with many people staying for a long time in a small area.
Such phenomena can partly be better represented by the discrete models or the homogeneous GP.
The advantage of heteroscedastic modeling of CoPA-Map becomes clear in the \emph{moving} case, where the method gives significantly better results.
Singularly occurring high target values (e.g. due to very short observation durations) are given less weight by CoPA-Map by adjusting the variance during training.
Fr-AAM, on the other hand, strongly approximates areas with high numbers of people, but as a discrete model suffers in terms of error metrics when people appear in slightly different locations in the test data.

The path with the largest area coverage at the smallest resolution (ATC, 0.5\,m $\times$ 30\,min) resulted in 147,500 input points and ca. 2860 inducing points. 
Training took a maximum of 28 minutes to converge in this case (\emph{Nvidia} GTX1070, i7-8700 CPU, 16 GB RAM). 
A duration of this magnitude thus makes it possible to repeat the training periodically (e.g. during the charging process) with current data.

\subsection{Service Disturbance }
\begin{figure}
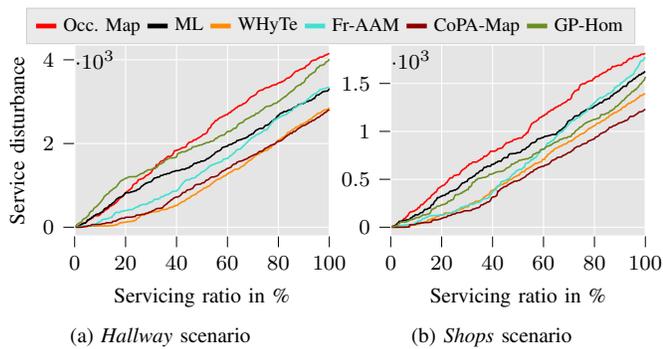

		\vspace{2mm}
	\noindent\makebox[\linewidth][c]{
		\begin{minipage}[t]{0.99\linewidth}
		\centering
		
		\resizebox{0.999\linewidth}{!}{
			
			\pgfplotsset{
	legend image code/.code={
		\draw[mark repeat=2,mark phase=2]
		plot coordinates {
			(0cm,0cm)
			(0.15cm,0cm)        
			(0.3cm,0cm)         
		};%
	}
}
\begin{tikzpicture} 
\definecolor{color0}{rgb}{1,0.549019607843137,0}
\definecolor{color1}{rgb}{0.250980392156863,0.87843137254902,0.815686274509804}
\definecolor{color2}{rgb}{0.419607843137255,0.556862745098039,0.137254901960784}
	\begin{axis}[%
		hide axis,
		width=\linewidth,
		xmin=10,
		xmax=12,
		ymin=0,
		ymax=0.4,
		legend cell align={left},
		legend columns=6,
		legend pos=north east,
		legend style={
			fill opacity=0.8,
			draw opacity=1,
			text opacity=1,
			at={(0,10)},
			draw=white!80!black,
			fill=white!89.8039215686275!black
		},
		legend style={font=\footnotesize},
		]
		\addlegendimage{line width=2pt, red}
		\addlegendentry{Occ. Map}
		\addlegendimage{line width=2pt, black}
		\addlegendentry{ML}
		\addlegendimage{line width=2pt, color0}
		\addlegendentry{WHyTe}
			\addlegendimage{line width=2pt, color1}
		\addlegendentry{Fr-AAM}
		\addlegendimage{line width=2pt, red!54.5098039215686!black}
		\addlegendentry{CoPA-Map}
		\addlegendimage{line width=2pt, color2}
		\addlegendentry{GP-Hom}

	\end{axis}
\hspace{-1mm}
\end{tikzpicture}
		}
	\end{minipage}}

		\begin{subfigure}[t]{0.49\linewidth}
		\setlength{\figureheight}{0.6\linewidth}
		\setlength{\figurewidth}{0.8\linewidth}
		\centering
		\input{fig/eval_soccost_gang.tex}
		\vspace{-5mm}
		\caption{\emph{Hallway} scenario}
		\label{fig:soccost_gang}
	\end{subfigure}
	\hfill
	\begin{subfigure}[t]{0.49\linewidth}
		\setlength{\figureheight}{0.6\linewidth}
		\setlength{\figurewidth}{0.8\linewidth}
		\centering
		\input{fig/eval_soccost_entry.tex}
			   \vspace{-5mm}
		\caption{\emph{Shops} scenario}
		\label{fig:soccost_entry}
	\end{subfigure}

	\caption{Service disturbance (encounters) for two navigation scenarios on the ATC dataset (lower is better). Smaller servicing ratios give more freedom to avoid peak hours of human activity and indicate if a model accurately captures temporal variations.}
			\label{fig:soccost}
		\vspace{-3mm}
\end{figure}

\begin{figure}
	\centering
	\includegraphics[width=1.0\linewidth]{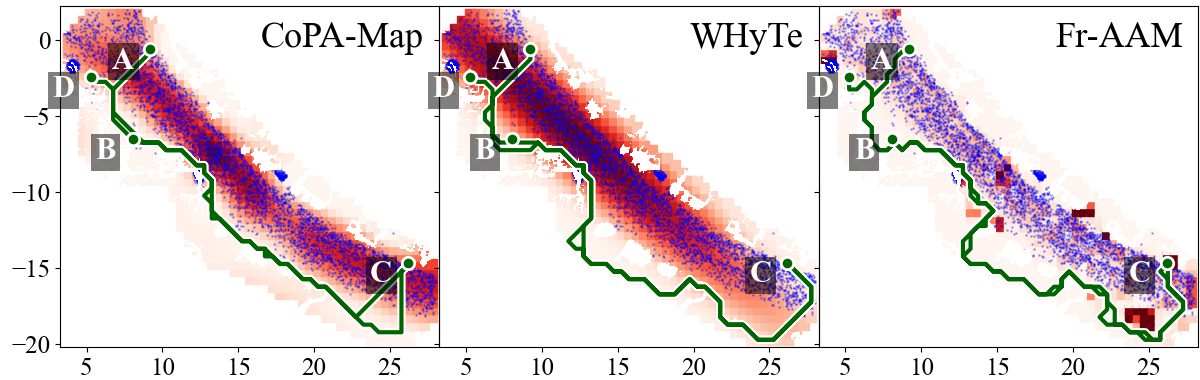}
	\includegraphics[width=1.0\linewidth]{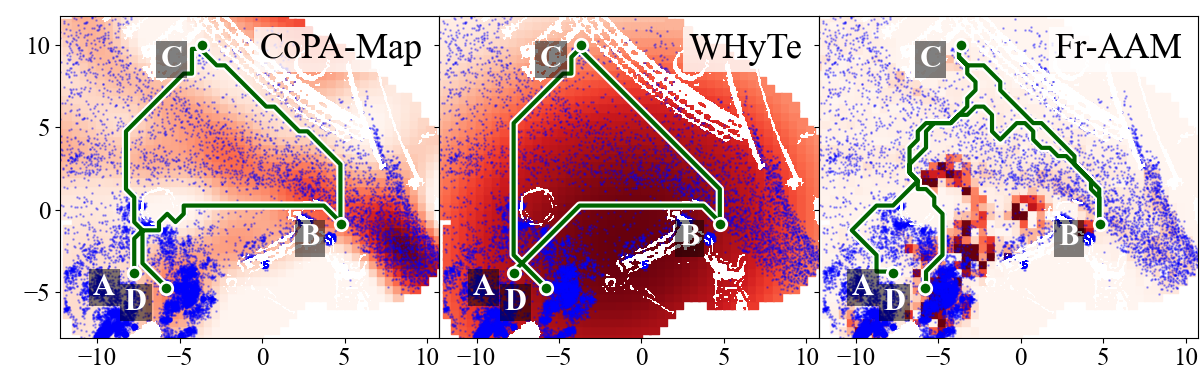}
	\caption{Exemplary model predictions (intensity of red color scaled to respective maximum model output) and resulting paths (green) from the service disturbance experiment. The upper images represent the \emph{hallway}, the lower images represent the \emph{shops} scenario. Pedestrian data is shown as blue dots. Obstacles and areas outside the \ac{fov} are masked in white. Models requiring a grid representation were trained with resolution 0.5\,m $\times$ 60\,min.}
	\label{fig:soccost_paths}
	\vspace{-5mm}
\end{figure}

The ability of the model to provide for active avoidance of areas with high human activity during navigation is also considered in two scenarios.
Both the \emph{hallway} and \emph{shops} scenarios are created based on the ATC dataset, the former involving a strong flow of people and the latter a splitting of people movement and longer stationary stays.
The input data was again created by a simulated robot movement and the edge weights of the resulting cost map are not directional.
Navigation scenarios are created between four positions (A$\shortto$B$\shortto$C$\shortto$D, depicted in Fig.~\ref{fig:soccost_paths}), five times per hour between 9 a.m. and 9 p.m. leading to $p=240$ scenarios for all four test days.
As a baseline, the \emph{Occupancy Map} model indicates how many encounters would occur, if the robot would always drive the shortest route by metric distance.
Fig. \ref{fig:soccost} shows the number of encounters (service disturbance) over the servicing ratio $r$ and Fig.~\ref{fig:soccost_paths} gives exemplary model outputs and navigational paths.
CoPA-Map and WHyTe as spatially continuous models capture the modality of human activity in the \emph{hallway} scenario significantly better than the discrete models.
These models, such as Fr-AAM, often lead to sinuous paths, increasing the number of human encounters.
CoPA-Map and WHyTe perform well for smaller service ratios ($<\SI{40}{\percent}$), since only the navigation tasks in the morning and evening hours are carried out, during which fewer people are expected.
Compared to WHyTe, CoPA-Map has advantages when multimodal pedestrian movements occur, as can be seen in the \emph{shops} scenario.
As WHyTe does incorporate the detections $\vec{p}_k$ directly, areas with shorter detection times, and thus fewer detections, might be underrepresented in the data.
The underlying \ac{gmm} is then more likely to underfit.
In contrast, e.g. in Fig.~\ref{fig:soccost_paths} (lower) CoPA-Map more accurately represents areas with many pedestrians, resulting in better paths for people avoidance.
For a servicing ratio of $r=1$ CoPA-Map leads to ca. \SI{32}{\percent} less encounters over all paths compared to the Occupancy Map model for both the \emph{hallway} and \emph{shops} scenarios.

	\section{Conclusion}
		\label{sec:conc}
We present CoPA-Map, a non-parametric method for spatio-temporal continuous modeling of human activity.
Compared to other methods, CoPA-Map has advantages with respect to the quality of predictions and path planning, especially when pedestrian data is collected by moving robots.
The model provides the basis for an extendable framework, that could also e.g. incorporate temporal trends through non-stationary covariance functions.
As CoPA-Map is a single-output model, extensions need to be investigated to also incorporate movement direction and speed of pedestrians, e.g. by multi-output Gaussian Processes.
	\bibliography{library}
	\bibliographystyle{IEEEtran}
	
	\end{document}